\documentclass[10pt,twocolumn,letterpaper]{article}

\usepackage{ijcb}
\usepackage{times}
\usepackage{epsfig}
\usepackage{graphicx}
\usepackage{amsmath}
\usepackage{amssymb}

\usepackage{times}
\usepackage{epsfig}
\usepackage{graphicx}
\usepackage{amsmath}
\usepackage{amssymb}
\usepackage{multirow}
\usepackage{bbm}

\usepackage{soul}

\ijcbfinalcopy 

\begin{document}

\title{CCFace: Classification Consistency for Low-Resolution Face Recognition}

\author{Mohammad Saeed Ebrahimi Saadabadi, Sahar Rahimi Malakshan,\\
Hossein Kashiani, and Nasser M. Nasrabadi\\
{\tt\small{me00018, sr00033, hk00014}@mix.wvu.edu, nasser.nasrabadi@mail.wvu.edu}
}

\maketitle
\thispagestyle{empty}
\pagestyle{empty}
\begin{abstract}
   In recent years, deep face recognition methods have demonstrated impressive results on in-the-wild datasets. However, these methods have shown a significant decline in performance when applied to real-world low-resolution benchmarks like TinyFace or SCFace. To address this challenge, we propose a novel classification consistency knowledge distillation approach that transfers the learned classifier from a high-resolution model to a low-resolution network. This approach helps in finding discriminative representations for low-resolution instances. To further improve the performance, we designed a knowledge distillation loss using the adaptive angular penalty inspired by the success of the popular angular margin loss function. The adaptive penalty reduces overfitting on low-resolution samples and alleviates the convergence issue of the model integrated with data augmentation. Additionally, we utilize an asymmetric cross-resolution learning approach based on the state-of-the-art semi-supervised representation learning paradigm to improve discriminability on low-resolution instances and prevent them from forming a cluster. Our proposed method outperforms state-of-the-art approaches on low-resolution benchmarks, with a three percent improvement on TinyFace while maintaining performance on high-resolution benchmarks.
\end{abstract}

\section{Introduction}
\begin{figure}[t]
\begin{center}

\includegraphics[width=1\linewidth]{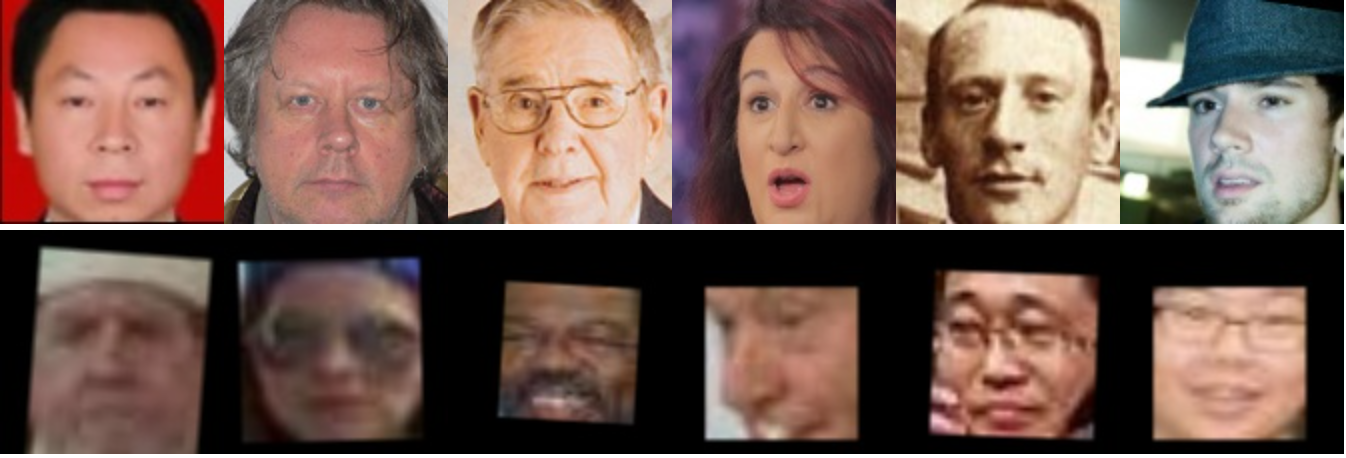}
\end{center}
\vspace{-3mm}
   \caption{\textbf{Top}: Samples from the training datasets are mainly high-quality images. \textbf{Bottom}: TinyFace contains real-world low-quality instances. Comparing the top with the bottom, the gap between training and low-resolution testing benchmarks is apparent.}\label{trainingvstinyface}

\end{figure}

One of the key factors of the recent advances in Face Recognition (FR) is the introduction of large-scale datasets \cite{deng2019arcface,liu2022sphereface,liu2017sphereface,liu2016large}.
Publicly available training benchmarks, such as CASIA-WebFace \cite{yi2014learning},  MS1MV2/3 \cite{guo2016ms, deng2019arcface}, and WebFace4M \cite{zhu2021webface260m}, are rich in width (thousands of identities) and depth (number of images per identities) \cite{du2020semi}. 
However, large-scale training datasets consist of web-crawled images and mostly contain high-resolution instances \cite{robbins2022effect}. 
As a result, there is a notable difference between training and real-world testing statistics  \cite{deng2023harnessing}. In particular, the images captured by security cameras exhibit a lower image resolution than the samples used for training, as seen in Fig. \ref{trainingvstinyface}
\cite{cheng2018low}.
This disparity leads to a huge performance gap in FR between High-Resolution (HR) and Low-Resolution (LR) since LR data are underrepresented during training \cite{Saadabadi_2023_WACV}. 
Biased training loss favors over-represented samples, hindering learning under-represented variations \cite{dehkordi2022multi}.
Therefore, studies have focused on developing better training objectives to solve this generalization problem \cite{saadabadi2022information,Saadabadi_2023_WACV}.

\begin{figure}[t]
\begin{center}
\includegraphics[width=0.95\linewidth]{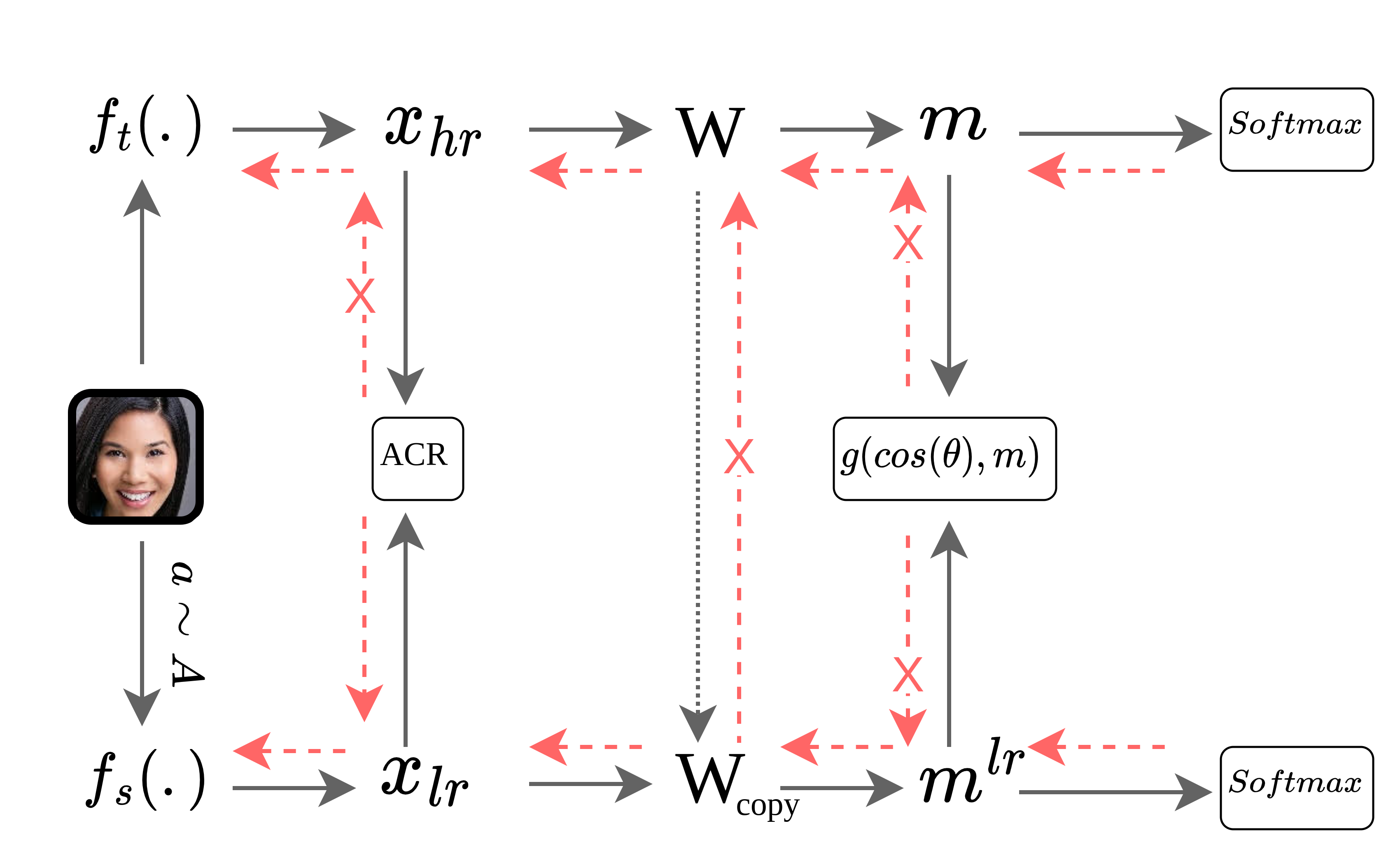}
\end{center}
   \caption{Our method aims to improve FR performance on LR input while maintaining the discriminability of the original HR embeddings. To achieve this, we propose to share the class proxies between student and teacher networks while asymmetrically pushing the LR feature embeddings to have higher mutual information with their HR counterparts.}
\label{long}
\label{fig:overview}
\end{figure}

Whether a set of samples or a proxy represent an identity, FR training criteria can be categorized as proxy-less or proxy-based methods \cite{du2020semi}.
The former is based on pair-wise similarities,  such as contrastive or triplet loss learning \cite{saadabadi2022information,schroff2015facenet}. In the latter, a prototype (proxy) represents a person's identity and the network tries to learn a classification task (weights of the classifier represent the identities' proxy).
In large-scale datasets, challenges concerning computationally expensive sample-mining of the proxy-less loss functions have led the current state-of-the-art (SOTA) FR training loss functions toward proxy-based approaches, such as L-Softmax \cite{liu2016large}, SphereFace \cite{liu2017sphereface}, CosFace
\cite{wang2018cosface}, and ArcFace \cite{deng2019arcface}.
Despite the remarkable improvement in numerous benchmarks, such as CFP-FP, LFW, CPLFW, and CALFW, significant performance degradation occurs when a FR is trained on these datasets and then applied to LR images \cite{shin2022teaching, Saadabadi_2023_WACV}.

Two primary approaches have been explored to combat this: 1) construction-based and 2) projection-based methods.
Construction-based methods involve enhancing the visual quality of the LR input before recognition, i.e., Face Super Resolution (FSR). This way, the FR process is separated into two tasks: identity-preserving FSR and Super Resolved Face Recognition (SRFR).
Among face generation modules, special attention has been placed on Generative Adversarial Networks (GANs) \cite{tran2017disentangled, ju2022complete,abbasian2023controlling}. Despite the remarkable outputs concerning image quality and human perception, GANs add high-frequency components to the synthesized images, which negatively affects the recognition process \cite{wang2018orthogonal}.
Furthermore, since multiple HR faces exist for each LR image, FSR is an ill-posed problem \cite{huang2017beyond}. Also, face images suffer from several other covariates (nuisance) factors, such as head pose, illumination, and expression. These factors result in a large gap between feature embeddings of HR and SR faces in the identity metric space, which significantly deteriorates the final FR performance \cite{wang2018orthogonal}.

Projection-based methods aim to create a shared embedding that can accommodate HR and LR images.
To this end, synthetic LR data can be used to increase the resolution diversity of the dataset \cite{shin2022teaching,kim2022adaface}. However, due to the fixed-angular margin in conventional FR methods, they suffer from convergence problems and cannot fit well with data augmentations such as down-sampling or random cropping \cite{zhang2023unifying}. To address this issue, methods have been proposed to adaptively tune the margin based on the difficulty of the samples \cite{liu2019adaptiveface,meng2021magface}. MagFace proposes to use the feature norm as the image-quality measure and tunes the margin. 
The adaptive margin has resolved the convergence problem to some extent. However, the performance still severely deteriorates when dealing with LR images \cite{Saadabadi_2023_WACV}. For instance, typically the face verification accuracy on LFW is above 99\%. However, the performance on Tinyface is around 59\%.
Furthermore, nourelahi et. al.  \cite{nourelahi2022explainable} demonstrate that training a model on the perturbed data costs worse performance on the original samples while increasing the robustness. 

Another line of work is to use Knowledge Distillation (KD) to obtain resolution agnostic face representation \cite{talreja2019attribute}. The main idea is to transfer the prior knowledge from HR images to train a model on the LR instances \cite{shin2022teaching}. 
A teacher network trained on HR images guides the LR model toward capturing discriminative features from LR instances. Since FR is an open-set problem, forcing the LR model to share the embedding space with the HR model is essential. A straightforward solution is to directly minimize the Euclidian distance between LR and HR model representations, which aligns the embedding spaces;
we call it Feature Distillation (FD). Previous methods mainly focused on FD because the embedding of the HR model includes more information than the LR model. In practice, it is shown that more than a FD is needed to align the models' representations \cite{shin2022teaching,Saadabadi_2023_WACV}. 
Enforcing a sample-level restriction on cross-resolution FR in a rigorous manner can be sub-optimal, as it eliminates the effect of negative instances and cause the network to prioritize factors such as the pose, glasses, or other facial attributes.

In this paper, we propose the Classification Consistency Face (CCFace) recognition paradigm, which takes advantage of KD and unsupervised representation learning to enforce a consistency between logits and feature embeddings of HR, and LR pairs, see Fig. \ref{long}. CCFace shares the same proxy between the HR and LR images and uses the output score of the HR images as a measure of sample hardness to tune the margin penalty of LR samples. In this manner, the training objective is relaxed for the LR inputs to alleviate the overfitting and convergence problem. As the HR embedding is more discriminative, we use two asymmetric methods to maintain the model's performance on HR images: 1) updating the proxies only from HR loss, 2) applying Asymmetric Cross-Resolution (ACR) learning between the HR and LR images with the detached features of HR inputs.
Contributions of this work can be summarized as follows:
\begin{itemize}
  \item We introduce a framework to maintain the consistency between HR and LR face recognition by sharing the proxies between different resolutions and asymmetrically updating the proxies via gradient from HR prediction.
  \item We introduce sample-mining in tuning the angular margin for LR samples to relax the training for hard samples and reduce overfitting.
  \item We asymmetrically apply contrastive learning to align the representation of HR and LR images without harming the discriminative power of the model on the original HR images.
\end{itemize}

\section{Related Works}

\subsection{General Face Recognition}\label{generalFR}
FR is among the oldest and most surfed problems in computer vision and has matured over the years from utilizing handcrafted features and local descriptors toward using deep learning-based models \cite{saadabadi2022information}. Existing deep network architectures, such as CNNs and ViT variants \cite{deng2019arcface,dosovitskiy2020image}, are dominant feature extractors in this area \cite{deng2019arcface,liu2022sphereface}. The critical issue is how to train the deep model using a large-scale dataset and prevent overfitting \cite{Saadabadi_2023_WACV}. 
Deep FR training schemes are either proxy-less or proxy-based. The former utilizes a tuple of similar and dissimilar images. It encourages the network to map the faces with the same identity to close representations and faces with distinct identities to distant representations \cite{saadabadi2022information}. The fact that these loss functions directly supervise sample-wise similarities is aligned with the final objective of the FR system. 
However, the sampling process required for these methods becomes challenging in large-scale datasets, which leads to convergence problems \cite{kim2020broadface}. Therefore, most of the research has been dedicated to classification-based loss functions, such as L-Softmax \cite{liu2016large}, SphereFace \cite{liu2017sphereface}, CosFace
\cite{wang2018cosface}, and ArcFace \cite{deng2019arcface}.

\subsection{Low-Resolution Face Recognition}

LR facial images lack the details of their HR counterpart, such as eyes and skin texture. Therefore, learning discriminative representations from LR samples are much harder, and applying general FR learning methods results in unsatisfactory performance \cite{shin2022teaching}. Two main approaches have been surfed for LR face recognition: 1) construction-based and 2) projection-based methods \cite{zhang2018super,wang2014comprehensive,cao2017attention,khosravi2022improved,shin2022teaching,ge2020look,zhu2019low}.

Construction-based methods are based on super-resolving the LR input prior to recognition. Apart from the ill-posed nature of the FSR problem, the main scheme of FSR is not optimized for discrimination, resulting in the unsatisfactory performance of the subsequent FR task \cite{saadabadi2022information}. To alleviate this, Zhang et al. \cite{zhang2018super} use the face verification loss in the identity metric space to guide the reconstruction model toward preserving identity information. 
Several studies focused on dividing the FSR into different sub-tasks. Wang et al. in \cite{wang2014comprehensive} dedicated one network to reconstructing global details and the other to enhancing local details. Cao et al. \cite{cao2017attention} used Reinforcement Learning (RL) to localize regions and a local network to attend to the specified regions. 
Current FSR methods produce visually appealing results; however, their integration with the FR model degrades the overall performance \cite{shin2022teaching}. Also, these methods are computationally intensive \cite{shin2022teaching}.

The projection-based approaches try to map both HR and LR images to a unified embedding \cite{ge2020look,zhu2019low}.
To this end, various works focused on distilling well-constructed features from HR teacher to LR student module \cite{zhu2019low}. Zhu et al. \cite{zhu2019low} utilized the general KD approach of using soft-logits prediction of the teacher module to guide the student module for the task of LR classification. Numerous studies applied intermediate representation distillation from the HR network to improve LR performance \cite{park2019relational,ge2020look}. However, the consistency between HR and LR representations was not imposed, deteriorating the performance on cross-resolution scenarios.
Among the facial parts, key parts are essential in FR, such as eyes and ears \cite{kumar2020s2ld}. Kumar et al. in \cite{kumar2020s2ld} force the model to generate key points using an auxiliary layer which guides the network toward focusing more on key facial characteristics. These methods mainly focused on improving the results of the LR images and did not maintain the performance of the original HR imagery. Also, guiding the LR network using the proxies of the HR network has not yet been explored for LR face recognition. Therefore, we investigate a prediction consistency knowledge distillation approach to improve the performance on LR images and preserve the model performance on HR images.

\subsection{Knowledge Distillation}
Knowledge distillation (KD), first proposed in \cite{hinton2015distilling}, is a form of model compression that transfers knowledge from a robust teacher model into a small student model. 
Since its introduction, numerous distillation techniques have been developed \cite{romero2014fitnets,tian2019contrastive,shin2022teaching}. 
FitNet \cite{romero2014fitnets} directly reduces the Euclidian distance between the student and teacher network. Combining representation learning with KD, CRD \cite{tian2019contrastive} utilizes contrastive objective to increase the mutual information between the teacher and student model. 
Despite the remarkable improvement in conventional KD, these methods are mainly designed for closed-set classification and are incompatible with FR. For instance, in \cite{hinton2015distilling}  Kullback-Leibler (KL) divergence between soft logits of teacher and student network is used to distill knowledge.
In an open-set problem such as FR, obtaining the discriminative feature is most important. Furthermore, consistency between features obtained from the teacher and student model is essential because cross-resolution embedding of a specific identity must be close to each other. 

\section{Method}
This section begins with a preliminary proxy-based FR loss function. Then, we explain our proposed classification consistency framework and how we prevent overfitting and alleviate the convergence problem. Then, we analyze the embedding of the FR module when dealing with LR instances and show that LR samples form a cluster well separated from other classes. Then, we introduce our asymmetric cross-resolution framework based on state-of-the-art representation learning methods to restrain the LR samples from forming a cluster. 

\subsection{Preliminary}\label{pre}

The majority of FR models consist of feature extractor $\mathcal{F}:\mathcal{I} \rightarrow \mathcal{X}$ mapping input faces $\mathcal{I}$ from image space to an embedding space $\mathcal{X}$. At the top of the feature extractor, there is a classifier $\mathcal{W}:\mathcal{X}\rightarrow\mathcal{\widehat{Y}}$ to predict the input identity from the embedding. Using gradient descent, both the feature extractor and classifier will be trained end-to-end. To this end, the widely used Softmax cross-entropy is applied on the predicted labels \cite{kashiani2022robust}:

\begin{equation}\label{Softmax}
 \small
 \begin{aligned}
		L_i= -log{\frac{e^{W_{y_i}^{T} x_i}}{\sum_{\substack{j=1}}^{C}e^{W_j^T x_i}}},
\end{aligned}
\end{equation}
where $\small{x_i \in \mathbb{R}^d}$ is the $\small{d}$-dimensional representation of $\small{i}$-th input. $\small{W \in \mathbb{R}^{d\times C}}$ represents the learnable matrix where each column $\small{W_j}$ is the proxy of $\small{j}$-th class. When both the $\small{x_i}$ and $\small{W_j}$ are mapped to a unit hypersphere, then the dot product $\small{W_j x_i}$ reflects the cosine of the angle between representation and proxy:

\begin{equation}\label{Softmax_angular}
 \small
 \begin{aligned}
		L_i= -log{\frac{e^{s \: cos(\theta_{{y_i}})}}{\sum_{\substack{j=1}}^{C}e^{s \: cos(\theta_{{j}})}}}.
\end{aligned}
\end{equation}

The introduction of angular penalty to the classification framework has been shown to be effective in increasing inter-class separability and intra-class compactness.  
SphereFace \cite{liu2017sphereface} argues that large-margin classification better aligns with open-set FR and introduces multiplicative angular margin for learning more discriminative features (period of the $cos(\theta_{y_i})$).
CosFace proposes using a vertical shift of the function $cos(\theta_{y_i})$, which leads to more powerful feature discrimination and improved stability. 
ArcFace suggests using additive angular margin (phase shift of $cos(\theta_{y_i})$), which has more clear geodesic interpretation and also improves the performance. In Eq.~\ref{combinedmarginloss}, $\small{m_s, m_c}$, and $\small{m_a}$ show the angular penalty introduced by SphereFace, CosFace, and ArcFace, respectively.  

\begin{equation}\label{combinedmarginloss}
 \small
 \begin{aligned}
 L_i= -log{\frac{e^{s \: cos({m_s}\theta_{y_{i}}+m_a)-m_c}}{e^{s \: cos({m_s}\theta_{y_{i}}+m_a)-m_c}+\sum_{\substack{j=1\\ j \neq y_i}}^{C}e^{s \: cos(\theta_{{j}})}}}.		
\end{aligned}
\end{equation}

\begin{figure}[t]
\begin{center}
\includegraphics[width=1.0\linewidth]{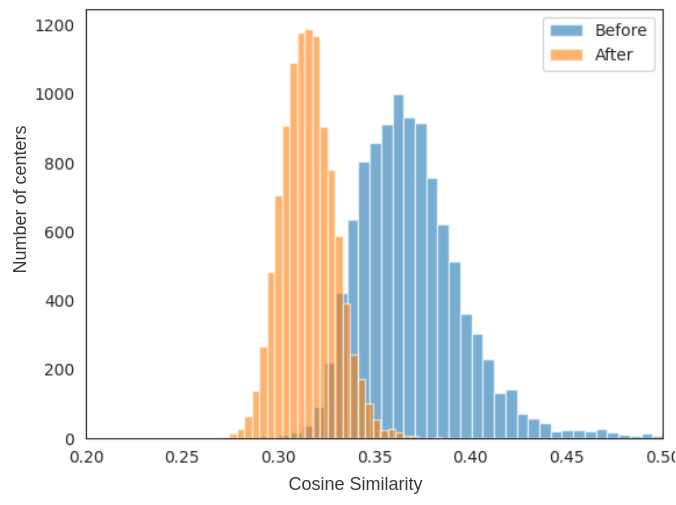}
\end{center}
   \caption{Maximum similarity between classifier proxies, before and after applying CCFace to a model trained by LR instances.}
\label{long}
\end{figure}

\subsection{Classification Consistency}
\begin{figure*}[t]
\begin{center}
\includegraphics[width=1.0\linewidth]{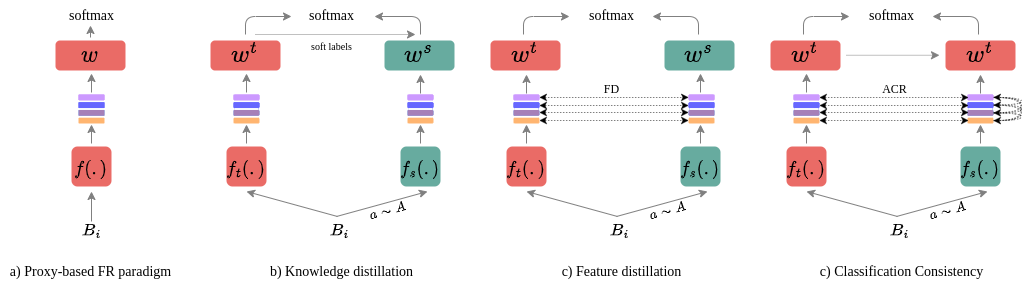}
\end{center}
   \caption{a) The general framework of proxy-based face recognition training has convergence issues when there are many hard samples (augmented data). b) The original knowledge-distillation paradigm in which the predicted probability of the teacher model is used as the target for the student model. c) Feature distillation derived from Knowledge-distillation. The similarity between teacher and student model features is being directly supervised only via positive pairs. d) Classification consistency framework in which the teacher model knowledge is transferred to the student model through the classifier proxies and the feature similarity supervision.}
\label{fig:long}
\label{fig:main}
\end{figure*}

Here unlike conventional KD approaches, the teacher and student networks have the same architecture. Instead, the teacher network only sees the original images, and the student network the down-sampled instances. 
Given a training set \( \small{ D}\), the LR samples are obtained from down-sampling with varying interpolations ($s$): \(\small{D_{LR}:{\{(I_{i}^{LR},y_{i}): I_i^{LR} = I_i \downarrow_{s} ) \}_{i=0}^{N}}, s \in S}\).
\(N\) is the total number of samples.
As presented in Fig.~\ref{fig:long}, \(\small{f_t(.)}\) and \(\small{f_s(.)}\), map the HR and LR faces to a \(\small{d}\)-dimensional embedding space; \(\small{x_i^{HR} = f_t(I_i) }\) and \(\small{x_i^{LR} = f_s(I_i^{LR})}\).

As discussed in section \ref{pre}, in the current SOTA FR objective functions, the similarity between features is guided through the cosine angle between features and softmax proxies. Therefore, the inter-class discrepancy increases if the proxies are well distributed on the hypersphere. Recently,  numerous studies have been conducted on uniformly distributing the proxies on the hypersphere \cite{liu2018learning}. In Fig. \ref{long}, we show our studies on the inter-class similarity from the proxies' viewpoint. We illustrate the maximum inter-class cosine value in two scenarios: 1) model trained on the original MS1MV2 dataset, 2) model trained on the down-sampled version of MS1MV2. This observation shows that naively training on the LR samples will result in poor discrimination (increase in inter-class similarity). The inter-class similarity is drastically increased in the model trained on LR images. Therefore, features being supervised through these proxies would not achieve preferred discriminability. 

To overcome this issue, we propose sharing the teacher network's proxies with the student model: $W^{HR}=W^{LR} = W$. However, the student model does not adjust the proxies and only uses them as a part of its forward propagation. Also, both networks are being trained using conventional large-angular margin loss function:

\begin{equation}\label{hr_angular}
 \small
 \begin{aligned}
		L_i^{HR}= -log{\frac{e^{s cos(\theta_{y_i}^{HR}+m)}}{e^{s \: cos({m_s}\theta_{y_{i}}+m)}+\sum_{\substack{j=1 \\ j \neq y_i}}^{C}e^{s cos(\theta_{j}^{HR})}}},
\end{aligned}
\end{equation}

\begin{equation}\label{lr_angular}
 \small
 \begin{aligned}
		L_i^{LR}= -log{\frac{e^{s cos(\theta_{y_i}^{LR}+m^{LR})}}{e^{s \: cos(\theta_{y_{i}}+m^{LR})}+\sum_{\substack{j=1 \\ j \neq y_i}}^{C}e^{s cos(\theta_{j}^{LR})}}},
\end{aligned}
\end{equation}
\noindent where $m$ and $m^{LR}$ are the angular margins applied to HR and LR samples, respectively. In this way, the feature representation of LR images is implicitly forced to move toward HR representation in high-dimensional embedding space. In the following section, we explain how we utilize the HR model prediction to tune the angular margin of the LR samples. Also, we further elucidate why simultaneously training two models helps the final performance of the student model.

\begin{figure}[t]
\begin{center}
\includegraphics[width=0.95\linewidth]{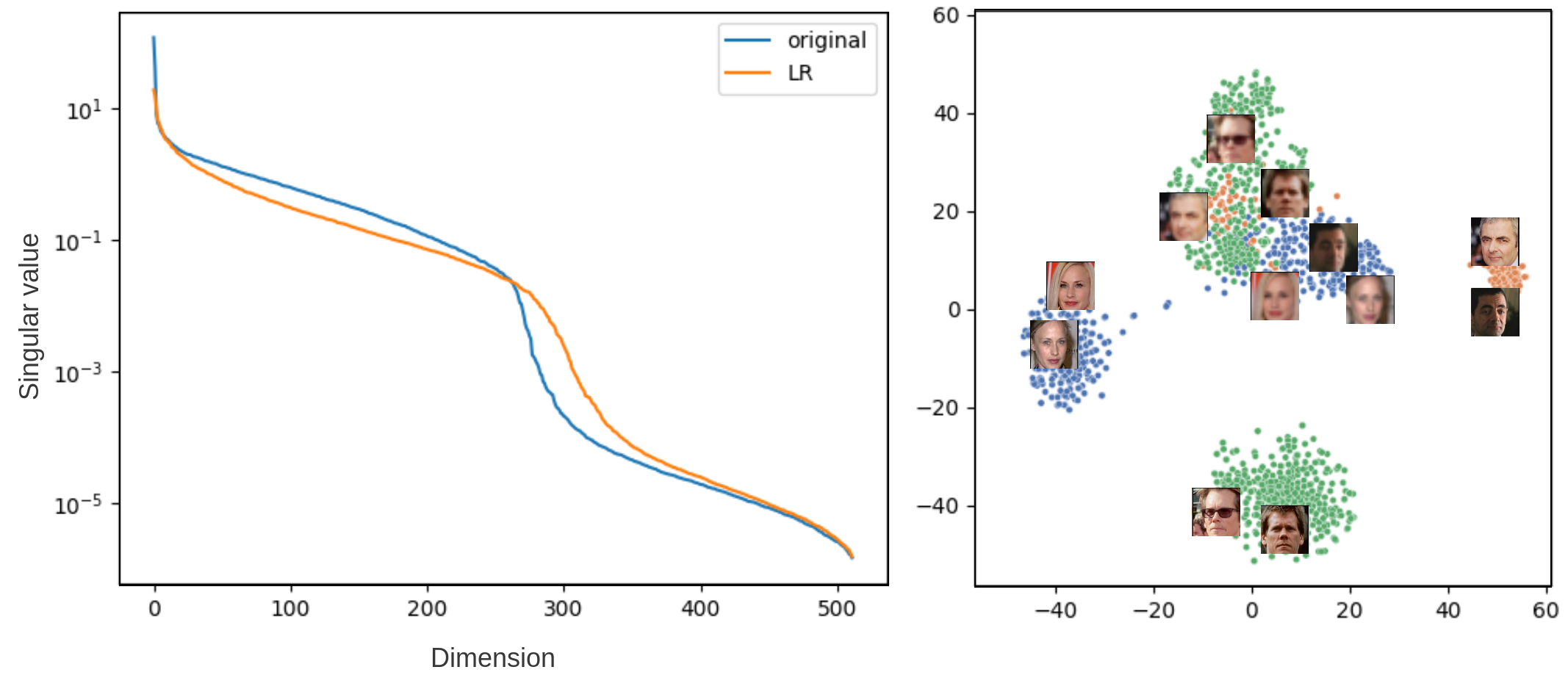}
\end{center}
   \caption{\textbf{left}: Singular value spectrum of embedding spaces. \textbf{right}: A visualization of hypersphere embeddings of the training dataset (every color represents an identity) generated by t-SNE [27].} \label{svd_tsne}


\end{figure}
\subsection{Cross-resolution Angular Margin Adaptivity}\label{adaptive_margin}
\begin{figure}[t]
\begin{center}
\includegraphics[width=1.0\linewidth]{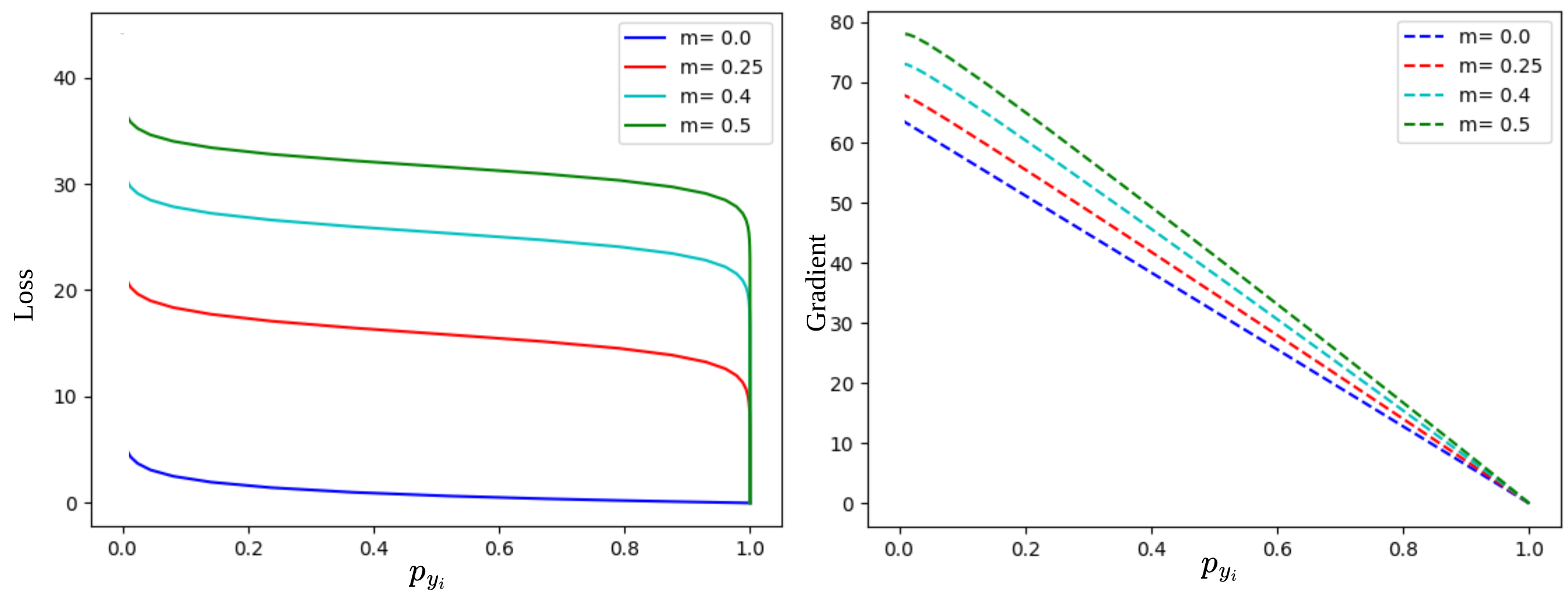}
\end{center}
\vspace{-3mm}
   \caption{Shows the curves for the loss values (left) and the gradient received to the backbone (right) versus $\small{p_{y_i} = \frac{e^{s(cos(\theta_{y_i}))}}{{\sum_{{j=1 }}^{C}{e^{s (cos(\theta_j))}}}}}$, when $cos(\theta_{y_i})$ changes from -1 to 1.}\label{lossVSgrad}

\end{figure}

Generally, there are three types of samples on the original training dataset: 1) easy, 2) hard, and 3) unrecognizable \cite{Saadabadi_2023_WACV}. Applying augmentation on different samples can produce different types of samples, whether unrecognizable or hard.
Angular margin helps the model push the features toward the corresponding positive proxy, i.e., by reducing the value of $cos(\theta)$. Larger margin results in more loss and eventually more gradient to adjust the network weights.
For a better illustration of the effect of angular margin on the loss and the magnitude of the gradient that the loss imposes on the network, we conduct a simple experiment. In a 2k classification problem, we changed the value of $cos(\theta_{y_i})$ from zero to one while the negative similarity scores with all of the classes are fixed to 0.1, $cos(\theta_j)=0.1;j\neq y_i$. Reducing the Eq.~\ref{combinedmarginloss} to ArcFace: 
\begin{equation}\label{arcfaceloss}
 \small
 \begin{aligned}
 L_i= -log{\frac{e^{s \: cos(\theta_{y_{i}}+m_a)}}{e^{s \: cos(\theta_{y_{i}}+m_a)}+\sum_{\substack{j=1\\ j \neq y_i}}^{C}e^{s \: cos(\theta_{{j}})}}},		
\end{aligned}
\end{equation}
by deriving the gradient of Eq.~\ref{arcfaceloss} with regard to $x_i$:
\begin{equation}\label{arcfacegrad}
 \small
 \begin{aligned}
 \frac{\partial L_i}{\partial x_i}&=  (p_{i,y_i}-1)\frac{\partial cos(\theta_{y_i}-m)}{\partial cos(\theta_{y_i})} w_{y_i}+\sum_{\substack{j=1 \\ j \neq y_i}}^{C}{p_{i,j}w_j} \\ & =\sum_{\substack{j=1}}^{C}{w_{j}}(p_{i,j}-\mathbbm{1}(y_i=j))(\frac{\partial cos(\theta_{j}-\mathbbm{m}(y_i=j))}{\partial cos(\theta_{j})}),		
\end{aligned}
\end{equation}

\begin{equation}\label{prob}
 \small
 \begin{aligned}
 p_{i,j}= \frac{exp(s\: cos(\theta_j + \mathbbm{m}(y_i=j)))}{exp(s \:cos(\theta_{y_i} + m))+\sum_{\substack{j=1\\ j \neq y_i}}^{C}e^{s \: cos(\theta_{{j}})}},	
\end{aligned}
\end{equation}
\noindent where we denote with $\mathbbm{m}[E]$ the indicator vector which outputs $m$ if the event $E$ is true and 0 otherwise. Note that the feature and softmax proxies are mapped to the unit hyper-sphere: $||w_j||=1$. Therefore, in Eq.~\ref{arcfacegrad} only the term $(p_{i,y_i}-\mathbbm{1}(y_i=j))(\frac{\partial cos(\theta_{j}-\mathbbm{m}(y_i=j))}{\partial cos(\theta_{j})})$ affects the gradient magnitude. In Fig. \ref{lossVSgrad}, we illustrate the gradient magnitude and the loss function value, which shows that the larger angular margin results in more adjustments for the backbone.
Therefore, applying the same angular margin on HR and LR samples can result in a convergence problem or overfitting of the student model \cite{kim2022adaface,meng2021magface,Saadabadi_2023_WACV}. We propose dynamically tuning the LR samples' margin value based on the HR samples' difficulty. One of the well-established sample difficulty measures is the softmax's output score. Here, we utilized the probability output of the teacher network to tune the margin value in the student model:
\begin{equation}\label{lr_angular}
 \small
 \begin{aligned}
		m^{LR}=max(0,m*cos(\theta_{y_i})),
\end{aligned}
\end{equation}
\noindent where $max(.)$ is for cases when the original sample is either unrecognizable or extremely hard for the teacher model. Therefore, applying margin on the LR version of that sample is not preferred \cite{Saadabadi_2023_WACV}. Since both models are being simultaneously trained, Eq.~\ref{lr_angular} helps the student model to ignore hard samples at the beginning of the training. Then at the final epochs, the model is able to concentrate more on the hard instances. 


\begin{table}[]
\addtolength{\tabcolsep}{7pt} 
\begin{center}
\small
\caption{Verification performance comparison on the IJB-B and IJB-C datasets.}
\vspace{1mm}
\centering
\small
\label{ijbbcresults}
\begin{tabular}{lcc}
\hline
\multirow{2}{*}{Method} & IJB-B            & IJB-C            \\ \cline{2-3} 
                        & \multicolumn{2}{c}{TAR@FAR: 0.0001} \\ \hline
CurricularFace          & 94.80            & 96.10            \\
MagFace                 & 94.51            & 95.97            \\
Mv-Arc-Softmax          & 93.6             & 95.2             \\ \hline
Ours                    & 94.91            & 96.29            \\ \hline
\end{tabular}
\end{center}
\end{table}

\subsection{Asymmetric Cross-Resolution Learning}
Here, we describe the Asymmetric Cross-Resolution Learning (ACR) learning of CCFace in detail. To begin with, we first discuss the need for the ACR learing. Then we describe our asymmetric loss function, which is based on the state-of-the-art approaches for representation learning.

During the training of angular-based FR algorithms, the network maps the hard instances of each class near the decision boundary \cite{meng2021magface}. The general idea is that the LR version of images would also be mapped near the class boundary since they are hard samples for the network \cite{deng2023harnessing}. However, studies have shown that the LR samples tend to cluster with each other \cite{robbins2022effect,deng2023harnessing}. Fig.~\ref{svd_tsne} shows this counterintuitive phenomenon, which we validate in the following experiment.
We experimented with the representation obtained from the original and down-sampled training instances. 
First, we randomly picked 1000 samples and computed their representations, $X \in R^{512\times 1000}$. Then, we compute the covariance matrix, $C \in R^{512\times 512}$:
\begin{equation}\label{cov}
 \small
 \begin{aligned}
    &C= \frac{1}{N}\sum_{i=1}^{N}{(x_i-\Bar{x}){(x_i-\Bar{x})}^{T}},
\end{aligned}
\end{equation}
\noindent where $N=1,000$ is the number of samples. Fig. \ref{svd_tsne} shows the singular value decomposition of $C$ in logarithmic scale and sorted order. As can be seen in this figure, the difference between the order of magnitude of singular values does not imply the dimensional collapse. From the observation in Fig. \ref{svd_tsne}, we can conclude that dimensional collapse is not happening in LR samples, and LR instances are forming a cluster well separated from the other classes. 
An explicit solution is reducing the pairwise distance between the normalized HR and LR representation of every subject presented in a mini-batch: 
\begin{equation}\label{contrastive}
 \small
 \begin{aligned}
	L_{fd} = \frac{1}{2N} \sum_{\substack{i=1}}^{N}{||\frac{x^{LR}_i}{x^{LR}_i}-\frac{x^{HR}_i}{x^{HR}_i}||^2},
\end{aligned}
\end{equation}
where $N$ is the number of samples in a mini-batch. $L_{fd}$ reduces the discrepancy between (HR, LR) pairs; However, rigidly adopting such a sample-level constraint to the cross-resolution FR is sub-optimal, i.e. no negative instances are in Eq. \ref{contrastive}. Using $L_{fd}$, the network may be induced to focus on the frontal pose, the glasses, or other non-identity facial attributes instead of the identity features of the HR-LR image pair.
To increase the cross-resolution intra-class compactness, the network should consider different samples of an identity. Furthermore, to prevent LR representations from forming a cluster, the loss function should increase the inter-class discrimination among LR representations. To this end, we proposed ACR to promote discrimination among the LR images and increase the mutual information among the HR and LR counterparts of the same subject: 
 \begin{equation}\label{cosineh}
 \small
 \begin{aligned}
	&h(x^{LR},x^{HR}) = \exp({\frac{1}{\tau}\frac{x^{LR} \cdot x^{HR}}{||x^{LR}|| \cdot ||x^{HR}||}}),
\end{aligned}
\end{equation}

\begin{equation}\label{supcon}
 \small
 \begin{aligned}
	&L_{acr,i}=-\log{\frac{1}{|P_i|}\sum_{p \in P_i}{ \frac{{h(x_{p}^{HR},x_{i}^{LR})}}{\sum_{j \in N_i}{{h(x_{j}^{LR},x_{i}^{LR})}}}}}.
\end{aligned}
\end{equation}
In Eq. \ref{supcon}, $N_i$ is a set of all negative LR samples presented in the mini-batch for $x_i$ ($N_i  \equiv {n \in B: y_n\neq y_i}$), $P_i$ is a set of all positive HR samples ($P_i  \equiv {p \in B: y_p=y_i}$) and $|P_i|$ is its cardinality. To maintain the performance on original HR data and since the HR representations have good intra-class compactness and inter-class discrimination, the detached HR features are used in Eq. \ref{supcon}. Therefore, Eq. \ref{supcon} asymmetrically forces the LR representation to increase their mutual information with their HR counterparts \cite{saadabadi2022information}. Also, it penalizes the similarity between negative LR representations, which prevents them from forming a cluster.
\section{Experiments}

\subsection{Datasets.} We report our experiments based on using MS1MV2 as the training dataset \cite{guo2016ms,deng2019arcface}.
The MS1MV2 dataset is a refined version of the MS-Celeb-1M dataset with 85k identities and 4 million images. Images were down-sampled using random interpolation to construct the HR and LR pairs. For evaluation, we report our results on LFW, CFP-FP, AgeDB, CPLFW, and CALFW. To validate the proposed method's performance on real-world LR faces, we also report our results on the TinyFace \cite{cheng2019low} and SCFace \cite{wallace2011inter} datasets.
We used an aligned version of these datasets in which samples are resized to 112 by 112 \cite{kim2022adaface}. Also, we report our performance on IJB-B and IJB-C as benchmarks which contain both HR and LR samples.

\textbf{IJB-B and IJB-C}: IJB-B \cite{whitelam2017iarpa} contains around 21.8K images (11.8K faces and 10K non-face images) and 7k videos (55K frames). A total of 1,845 identities are presented in this dataset. Our experimental protocols follow the standard 1:1 verification, which contains 10,270 positive and 8M negative matches. There are 12,115 templates in the protocol, each of which consists of multiple images or frames. Consequently, a template-based matching process is used. Specifically, we average over the instances in a template to obtain the global feature vector for each template. IJB-C \cite{maze2018iarpa} is the extended version of IJB-B, including 31.3K images and 117.5K frames from 3,531 identities. The testing protocol of IJB-C is similar to IJB-B.
\begin{figure}[t]
\begin{center}
\includegraphics[width=1.0\linewidth]{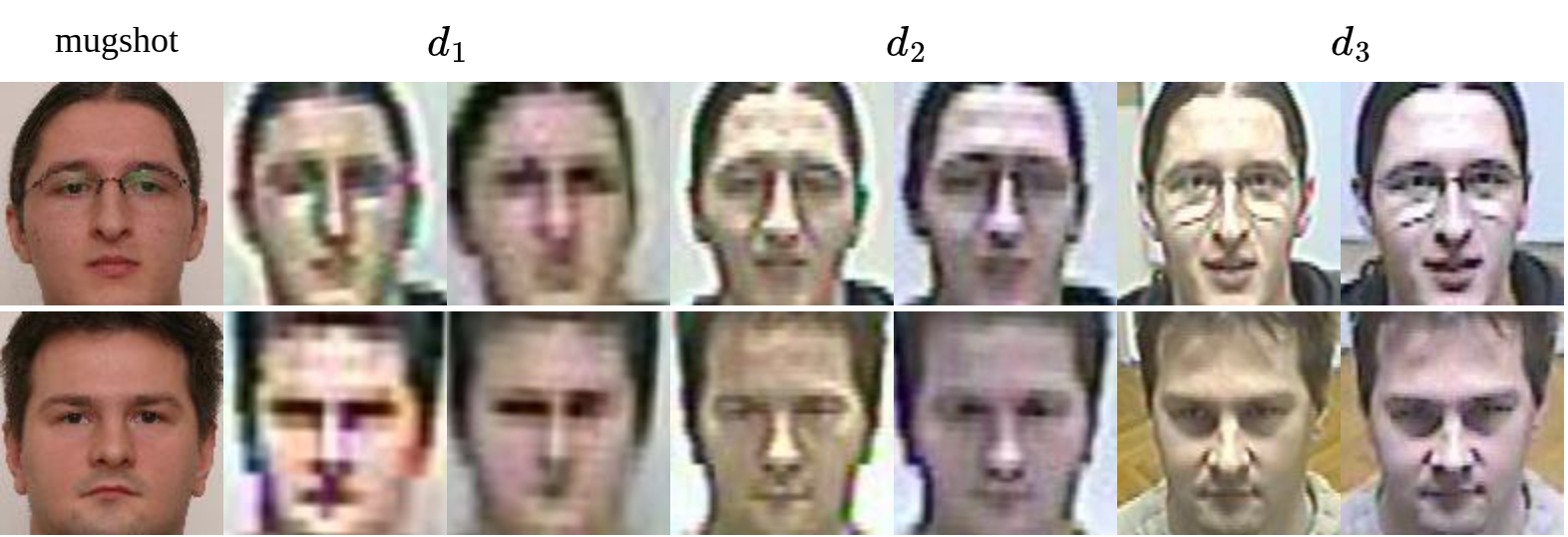}
\end{center}
\vspace{-3mm}
   \caption{Samples of SCFace. The first column shows the high-quality mugshot instances. For every distance $\{ d_1,d_2,d_3\}$, images are taken using seven different cameras.}
\label{typesOfSamples}
\end{figure}

\textbf{SCFace}: 
SCface [16] is a challenging cross-resolution FR dataset. It contains HQ mugshot and LQ images captured by surveillance cameras.
The images were taken from 130 subjects in an uncontrolled indoor environment using five video surveillance cameras at different distances $d_i \in \{4.2, 2.6, 1.0\}$ (meter); five images at each distance. Also,  one frontal mugshot image for each subject 
is obtained using a digital camera, Fig. \ref{typesOfSamples}. In an experiment similar to [16], we employ frontal mugshot images as our gallery and samples taken by surveillance cameras as probes.

\subsection{Metrics.} There are two primary approaches to validating an FR paradigm's performance: 1) Recognition and 2) Verification. Recognition is a 1:N task where the network should calculate the similarity score of a given probe image with all the samples in the gallery and determine the identity of the probe image. Face verification is a 1:1 task where the network should determine whether a given pair of images represent the same identity. We report the verification results on the  LFW, CFP-FP, AgeDB, CPLFW, IJB-B, IJB-C, and CALFW datasets (TAR@FAR from ROC for IJB-B and IJB-C \cite{ghavidel2023ensemble}). The result for identification is reported on the SCFace and TinyFace datasets. 

\subsection{Implementation details.} We followed the ArcFace setup for preprocessing \cite{deng2019arcface}. All the images are resized to 112\(\times\)112, aligned to a canonical view, and pixel values are normalized to \([-1,1]\). To produce synthetic low-resolution samples, we randomly apply different down-sampling interpolations on the $I_{HR}$, including bilinear, nearest neighbor, and bicubic. Also, the size of the LR image is uniformly sampled from $\{(16\times16), (20\times20), (32\times32)\}$. Together, there would be twenty-seven distinct augmentations. 
The experiments are conducted with ResNet100 as the backbone \cite{deng2019arcface,safavigerdini2023predicting} unless mentioned. The model is trained for 24 epochs with the Arcface loss. The optimizer is SGD, with the learning rate starting from 0.1 decreased by a factor of 10 at epochs \{10, 16, 22\}. The optimizer weight-decay is set to 0.0005, and the momentum is 0.9. During training, the mini-batch size on each GPU is 512, and the model is trained using two Quadro RTX 8000. After training is finished, we disregard the teacher model and only use the student model for evaluation.  

\begin{table}[]
\begin{center}
\addtolength{\tabcolsep}{-3pt} 
\small
\caption{Identification accuracy on TinyFace}
\centering

\label{TinyFace}
\begin{tabular}{lllll}
\hline
                & Architecture & DataSet & Acc@1                                                 & Acc@5          \\ \hline
AT \cite{zagoruyko2016paying} & ResNet50     & CASIA   & 36.54                                                 & 50.62          \\
HORKD \cite{ge2020look}        & ResNet50     & CASIA   & 45.49                                                 & 54.80          \\
A-SKD \cite{shin2022teaching}      & ResNet50     & CASIA   & 47.91                                                 & 56.55          \\  \hline
URL \cite{shi2020towards}   & ResNet100    & MS1MV2  & 63.89                                                 & 68.67          \\
CurricularFace \cite{huang2020curricularface}   & ResNet100    & MS1MV2  & 63.68                                                 & 67.65          \\  \hline
CCFace          & ResNet100    & MS1MV2  & \textbf{65.71} & \textbf{69.25} \\ \hline
\end{tabular}
\end{center}
\end{table}

\begin{table}[]
\begin{center}
\addtolength{\tabcolsep}{2pt} 
\caption{Identification accuracy on SCFace}
\label{SCFace}
\begin{tabular}{lccc}
\hline
\multirow{2}{*}{Method}                             & \multicolumn{3}{c}{Distance} \\ \cline{2-4} 
                                                    & d1       & d2      & d3      \\ \hline

SKD \cite{ge2018low}               & 43.5     & 48.0    & 53.50   \\
CGAN \cite{talreja2019attribute}   & 44.81    & 49.61   & 54.30   \\
MDS \cite{mudunuri2015low}         & 60.3     & 66.0    & 69.5    \\
DMDS \cite{yang2017discriminative} & 61.5     & 67.2    & 62.9    \\
VGGFace  \cite{parkhi2015deep}                                           & 41.3     & 75.5    & 88.8    \\
DCR \cite{lu2018deep}              & 73.3     & 93.5    & 98.0    \\
CCFace (r50,MS1MV2)                                                &  {74.8}        &  94.01       &  99.47       \\ \hline
\end{tabular}
\end{center}
\vspace{-5mm}
\end{table}

\subsection{Performance Comparison}
In this section, we assess CCFace's performance against the SOTA methods,  such as CurricularFace, MagFace, and URL, using TinyFace, SCFace, IJB-B, and IJB-C datasets. Results in Table \ref{ijbbcresults} demonstrate CCFace's competitive performance against other methods. Showing the efficacy of alignment between teacher and student model. Since IJB-B and IJB-C contain both HR and LR samples, this performance shows the ability of the proposed method to generalize across a wide range of resolutions. Table \ref{TinyFace} demonstrates the results on TinyFace. According to this result, CCFace can effectively boost the discrimination power over the LR data, which results in over 3 percent improvement in TinyFace data. To evaluate the performance of CCFace for cross-resolution scenarios, Table \ref{SCFace} shows the identification result on the SCFace dataset. CCFace could improve the results of the previous methods by more than one percent. Success across datasets with varying quality levels, from LR to HR, shows the efficacy in alignment between teacher and student embedding space, and maintaining the discriminability on both HR and LR images.

\section{Ablation Study}
\subsection{Impact of Augmentation}
One of the major issues with angular-margin-based loss functions, which are dominant in FR, is that their integration with data augmentation is challenging \cite{Saadabadi_2023_WACV}. In order to alleviate this issue, we use the output probability of the teacher network to tune the margin value for augmented samples; see section \ref{adaptive_margin} for more detail. Table. \ref{table4} shows the student model's performance with different augmentation probabilities. Increasing the augmentation occurrence boosts the FR performance on the LR faces considerably; however, negligible performance degradation is observed on high-quality benchmarks. Table. \ref{table4} also shows the performance of the student model without using an adaptive margin. Since the augmentation process increases the chance of unrecognizable instances, the performance gain is inconsistent. 

\subsection{Components of Proposed Training}
Here, we analyze the impact of different elements within our training paradigm on performance. Our experiments utilize R18 as the backbone and CASIA as the training dataset. As shown in Table. 5, training a network solely with a single classifier results in poor performance on the LR samples. However, sharing the classifier between the teacher and student improves performance to some extent. Additionally, incorporating an adaptive margin into the model enhances performance on the LQ dataset. Notably, the application of ACR between the teacher and student significantly improves the cross-quality scenario. 

\begin{figure}[t]
\begin{center}
\includegraphics[width=0.8\linewidth]{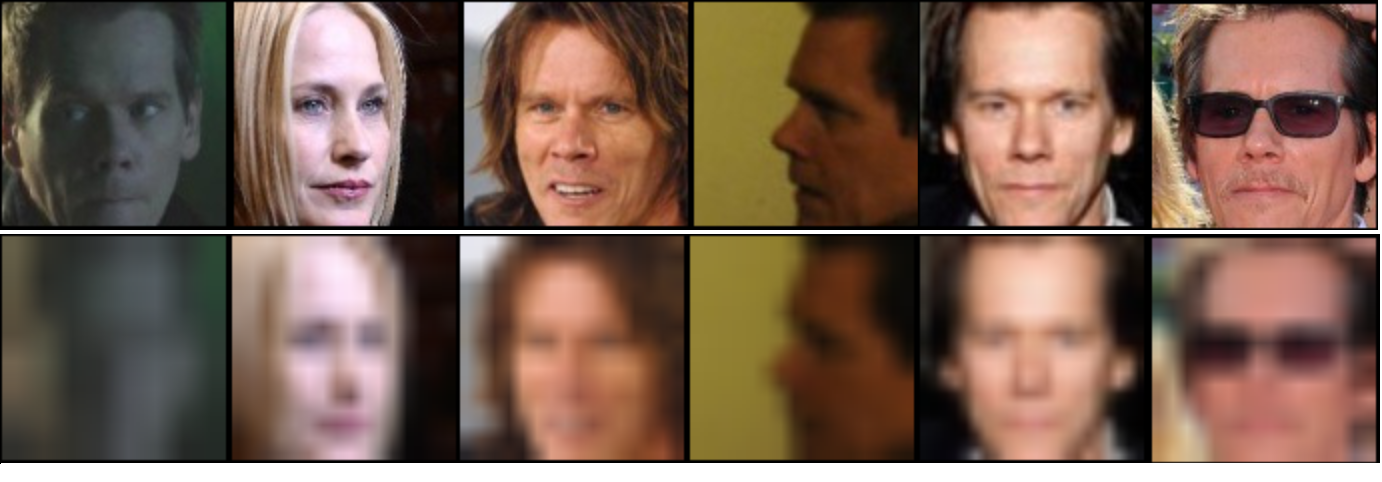}
\end{center}
\vspace{-3mm}
   \caption{\textbf{Top}: Original samples of the training dataset. \textbf{Bottom}: The augmented version of training samples.}
\label{fig:typesOfSamples}

\end{figure}

\begin{table}[]

\addtolength{\tabcolsep}{-3pt} 
\small
\caption{Verification TAR@FAR:0.0001 for the IJB-B and IJB-C datasets with different amounts of augmentation during the training.}\label{table4}
\begin{center}
\begin{tabular}{cccc}
\hline
Augmentation Prob. & Adaptive Margin & IJB-B & IJB-C \\ \hline
0.0        & \checkmark             & 94.53      & 95.29      \\
0.1        & \checkmark             &  94.74     & 95.59      \\
0.2        & \checkmark             &   94.91    &  96.29     \\ \hline
0.2        &               & 94.56      &  95.42     \\ \hline
\end{tabular}

\end{center}
\vspace{-5mm}
\end{table}

\addtolength{\tabcolsep}{-3pt} 
\begin{table}[t]
\small
\caption{Ablation study on the effect of the proposed method on the performance. CR: Cross-resolution matching (every testing pair contain one HR and one LR image), SC: Shared classifier, AM: Adaptive angular margin.  }
\vspace{-2mm}
\begin{center}
\begin{tabular}{cccccccccc}
\hline
\scalebox{.8}{Res}                      & \scalebox{.8}{CR}  & \scalebox{.8}{SC}  & \scalebox{.8}{AM} & \scalebox{.8}{ACR}  & \scalebox{.8}{LFW} & \scalebox{.8}{CFP-FP} & \scalebox{.8}{CPLFW} & \scalebox{.8}{CALFW} & \scalebox{.8}{Age-DB} \\ \hline
\multirow{4}{*}{\scalebox{.8}{16}}& \scalebox{.8}{}& \scalebox{.8}{\checkmark} & \scalebox{.8}{}& \scalebox{.8}{} &\scalebox{.8}{91.2} & \scalebox{.8}{73.3}   & \scalebox{.8}{74.3}  & \scalebox{.8}{75.3}  & \scalebox{.8}{68.7}        \\
                         &\scalebox{.8}{} & \scalebox{.8}{\checkmark} & \scalebox{.8}{\checkmark}& \scalebox{.8}{}&\scalebox{.8}{92.3}&\scalebox{.8}{75.9}&\scalebox{.8}{79.9}&\scalebox{.8}{78.4}&\scalebox{.8}{71.5}        \\
                         & \scalebox{.8}{}& \scalebox{.8}{\checkmark} & \scalebox{.8}{\checkmark}& \scalebox{.8}{\checkmark}&\scalebox{.8}{95.9} & \scalebox{.8}{83.6} & \scalebox{.8}{83.0}  & \scalebox{.8}{82.0}  & \scalebox{.8}{75.4}         \\
                         & \scalebox{.8}{\checkmark}& \scalebox{.8}{\checkmark} & \scalebox{.8}{\checkmark}& \scalebox{.8}{\checkmark}&\scalebox{.8}{92.1}& \scalebox{.8}{73.9}&\scalebox{.8}{80.1}&\scalebox{.8}{74.2}&\scalebox{.8}{68.8}        \\ \hline
\multirow{4}{*}{\scalebox{.8}{64}} & \scalebox{.8}{}& \scalebox{.8}{\checkmark} &\scalebox{.8}{}&\scalebox{.8}{}& \scalebox{.8}{99.7} & \scalebox{.8}{98.3}   & \scalebox{.8}{93.4}  & \scalebox{.8}{95.9}  & \scalebox{.8}{97.9}        \\
                         & \scalebox{.8}{}& \scalebox{.8}{\checkmark}& \scalebox{.8}{\checkmark}& \scalebox{.8}{}&\scalebox{.8}{99.7}&\scalebox{.8}{98.5}&\scalebox{.8}{93.3}&\scalebox{.8}{96.01}&\scalebox{.8}{98.12}        \\
                         & \scalebox{.8}{}& \scalebox{.8}{\checkmark} & \scalebox{.8}{\checkmark}& \scalebox{.8}{\checkmark}& \scalebox{.8}{99.8} & \scalebox{.8}{98.8}& \scalebox{.8}{93.3} & \scalebox{.8}{96.13} & \scalebox{.8}{98.3} \\
                         & \scalebox{.8}{\checkmark}& \scalebox{.8}{\checkmark} & \scalebox{.8}{\checkmark}& \scalebox{.8}{\checkmark}&\scalebox{.8}{99.7}&\scalebox{.8}{98.0}&\scalebox{.8}{94.4}&\scalebox{.8}{95.6}&\scalebox{.8}{97.9}        \\ \hline
\end{tabular}
\end{center}

\label{table:results_cmu}
\vspace{-6mm}
\end{table}

\section{Conclusion}
The proposed CCFace framework strives to maintain its performance on a wide variety of resolutions and generalize well for cross-resolution FR by simply sharing the classifier between the student and teacher network. We established that identities' proxies constructed from HR images are simple yet effective knowledge that can help the model to compensate for the information loss in the LR images and find a discriminative representation for the LR instances. Furthermore, our method prevents LR samples from forming a distinct cluster by applying pushing force directly between LR samples in a contrastive manner.

{\small
\bibliographystyle{ieee}
\bibliography{egbib}
}

\end{document}